\RequirePackage{amsmath} 
\documentclass[runningheads]{llncs}
\usepackage{graphicx}
\usepackage{paralist}   
\usepackage{comment}    
\usepackage{tikz-cd}    
\usepackage{subfig}     
\usepackage{eurosym}    

\begin{document}
\title{Value alignment: a formal approach}
%
%
\author{Carles Sierra \and
Nardine Osman \and
Pablo Noriega \and
Jordi Sabater-Mir \and
Antoni Perello-Moragues}
\authorrunning{C. Sierra et al.}
%
\institute{Artificial Intelligence Research Institute (IIIA-CSIC), Bellaterra, Catalonia}
\maketitle              
\begin{abstract}
Value alignment in AI has emerged as one of the basic principles that should govern autonomous AI systems. It essentially states that a system's goals and behaviour should be aligned with human values. But how to ensure value alignment? In this paper we first provide a formal model to represent values through preferences and ways to compute value aggregations; i.e. preferences with respect to a group of agents and/or preferences with respect to sets of values. Value alignment is then defined, and computed, for a given norm with respect to a given value through the increase/decrease that it results in the preferences of future states of the world. We focus on norms as it is norms that govern behaviour, and as such, the alignment of a given system with a given value will be dictated by the norms the system follows. 

\end{abstract}
\section{Introduction}




The aim of this paper is to explore the alignment of a given system with a given value. To achieve this, we explore the 
relationships between values, actions and norms in order to propose a precise way of expressing when a norm fosters behaviour in accordance to a value.  This is a first step towards a formal foundation for a theory of value-driven behaviour of autonomous entities, that is good enough to engineer value-imbued socio-cognitive technical systems.

Our proposal is based on four  main assumptions: 
First, we adopt a \textit{cognitive view of values}. That is, we understand values as a cognitive construct that is involved in the rational behaviour of an agent~\cite{Miceli1989-MICACA,Rohan2000,Parks2009,schwartz1992universals}. In line with Schwartz theory of motivational values~\cite{schwartz1992universals} we will assume that values serve as standards, refer to desirable goals and transcend specific actions.  


Our second assumption is a \textit{consequentalist} view of values ~\cite{sep-consequentialism} by which the ``worthiness'' of a value, and therefore, its social meaning, is given by the outcomes of the actions that are aligned with it. 

These two views allow us to postulate that values serve two decision-making purposes:
\begin{inparaenum}[(i)]
    \item to assess the ``worthiness'' of a state of the world (and thus compare the ``worthiness'' of two states of the world), and 
    \item to decide which is the ``better'' of two actions.
\end{inparaenum}

Next, we assume the usual \textit{teleological view of norms} that conceives norms as a social means to promote desired behaviours and to discourage undesired ones.  We follow the normative notions that are prevalent in the field of multiagent systems (f.e.  ~\cite{Andrighetto2013}). 
More specifically, in this paper we will assume that norms modify the outcomes of actions, by inducing  agent behaviour through the use of positive and negative incentives.
 The fourth assumption is to characterise the space where actions take place to be a situated, online, regulated open multiagent system, in fact a socio-cognitive technical system (SCTS)~\cite{NoriegaVdP16}.
 
Based on these assumptions, we make precise what it means for a norm to be aligned with a value. This is the basis for determining whether a system fosters behaviour that is in accordance with a value or not. 

To achieve that characterisation of value-alignment we first establish a relationship between values and preferences and apply this relationship to the evolving states of the world. After that, we discuss how norms are linked with actions and how to determine the value-related effects of performing a norm compliant action. With these elements we then define the key notion of norm-alignment. We illustrates these ideas by instantiating and programming them for a version of the prisoner's dilemma. The last section suggests some open problems.

\label{sec:motivation}

\section{Background}
\subsection{The Value-Alignment Problem (VAP)} 
The VAP is motivated by the recognition that autonomous entities (robots, software agents) and intelligent systems in general exhibit increasingly complex behaviour and thus become difficult to regulate.  One way to address this concern is to imbue values in intelligent systems. There are currently three main approaches to towards this aim. First the design of guidelines, standards and certifications~\cite{rizzo_ethically_2017}. Second, ``value-based design'' where values are brought into the systems from the very start as design requirements~\cite{vandePoel2013,VandePoel2009,rahwan}. The third approach postulates that autonomous entities should be provably made to comply with values~\cite{russell2017provably}. Our proposal is framed in this analytic approach. 
We follow a version of this approach framing the compliance problem within hybrid on-line social coordination systems (or \textit{socio-cognitive technical systems}) whereas autonomous rational entities interact within a norm-regulated shared social space~\cite{Aldewereld2016,NoriegaVdP16}. More specifically, we frame the problem of imbuing values in such systems by having norms that are aligned with some values and autonomous agents that, subject to the norms of a social coordination environment, may act in accordance to their own ---possibly different--- values.   

\subsection{Values}
\begin{figure*}
    \centering
    \includegraphics[width=\textwidth]{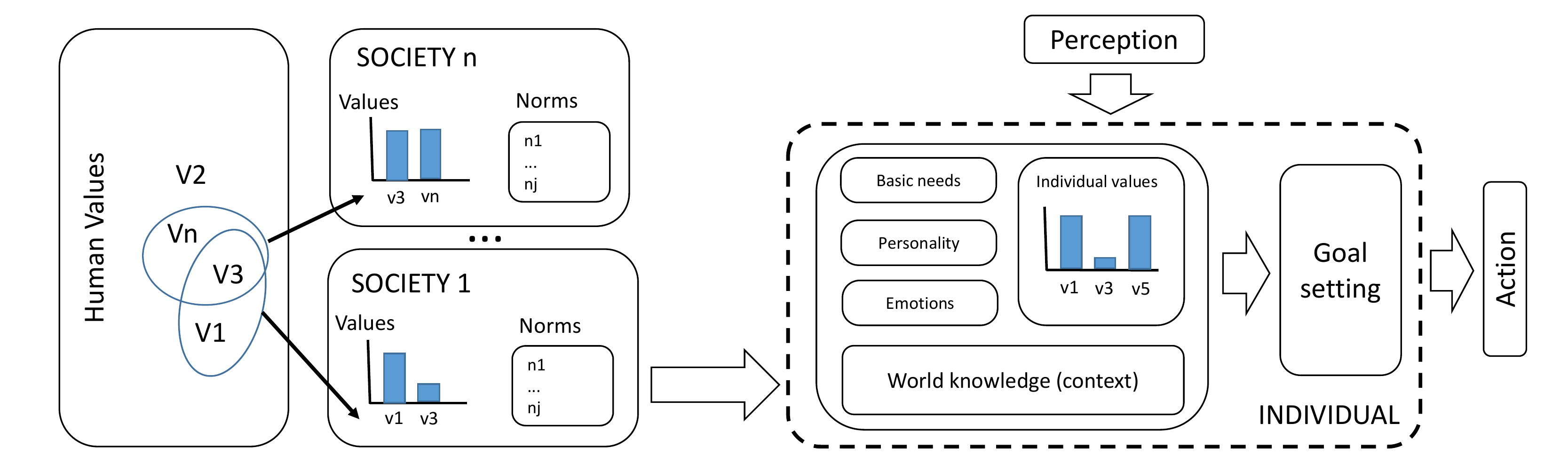}
    \caption{General view of an agent's value-guided behaviour}
    \label{fig:schema}
\end{figure*}

We want a notion of value that is linked to agent value-guided behaviour that presumes an empirically-grounded, motivational understanding of values
~\cite{Miceli1989-MICACA,Parks2009,Reiss2001,Mercuur2017,schwartz1992universals}. 
Drawing on Schwartz' theory of basic human values~\cite{schwartz1992universals}, we presume that each society adopts a finite set of basic human values, orders them by importance, and designs norms that foster behaviour that aligns with those values. Likewise, individuals also adopt and order a finite set of basic values that align with their own behavioural profile, and are influenced by social values and the corresponding norms. Individuals' goal setting depends on the context (which includes applicable norms) and the individuals' mind-frames (which include the \emph{individuals' values}, needs, personality, emotions and beliefs among other constructs). A full description of our proposal is beyond the scope of this paper but Figure~\ref{fig:schema} sketches our understanding of value-guided behaviour.

 
With this view we make values serve two main functions: to assess the ``goodness'' of the states of the world ---thus determining preferences between two states--- and to decide whether one action is preferable to another. 
In order to make these intuitions operational it is convenient to take a \textit{consequentialist} view of values by which value is reified by its consequences ~\cite{sep-consequentialism,vandePoel2013}. By so doing, we may use a representation of the relevant aspects of the world to measure the level of ``goodness''  of a given state of world. By the same token we may measure the goodness of a given state of the world and compare it with the goodness of the new state that is reached after a given action is taken.\footnote{Consequentalism eventually commits to commensurable values, however one needs not commit to any particular aggregation function. For a single system there may be different ways of computing the aggregation, different observable variables involved in a value and different forms to ``score'' those variables.}

\subsection{The Space of Action}

We follow the postulates of~\cite{NoriegaVdP16}:  (i) there are two primitive (different) components in a SCTS: social world and agents; (ii) the social world has a fixed ontology; (iii) at any moment in time there is an explicit \textit{state of the world} that is unique and the same for all agents; and (iv) the state of the world changes only through admissible events and those agent actions that comply with the regulations of the space. Furthermore, because of consequentialism, we assume (i) that the state of the world can be assessed ---and thus we can compare two states and hence prefer one over the other--- and (ii) that since the world changes when an agent executes an action, one may assess whether the effect  of  an action produces a better state of the world or not.


%
%
%

\subsection{Related Work}
%
In~\cite{Cranefield2017}, the authors use values to select among different possible plans. They use the same approach as~\cite{DiTosto2012}, where the agent has a desired level of satisfaction for each specific value. Different actions have different effects on the current satisfaction levels. 
For a given goal, the plan composed by actions that ``best'' modify the current value levels in the direction of the desired ones is the one that is preferred. Unlike these authors,  in this paper we are not concerned about the actual motivation of agents, nor about their evaluation process. However we share with them the notion that actions promote or demote values and that a hierarchy of values ---thus a notion of preference--- is one of the elements involved in the decision of which action to take.


Also close to our main assumptions are value-based argumentation proposals where the main idea is that an argumentation move will promote, demote or be indifferent towards a value~\cite{doi:10.1093/logcom/13.3.429}. In particular, 
~\cite{AtkinsonBM066} proposes an argument scheme (PRAS) where a transition between states of the world is labelled for each value 
according to the promotion/demotion effect induced by an action; sort of like we do but with the purpose that plans, goals and the promotion of values can be explicitly reasoned about. Likewise, in~\cite{vanderWeide2009}, the authors propose an argumentation mechanism (to deal with disagreement about the meaning of a value), where values are defined as preference orders, similar to our own, and the authors discuss the problem of arguing about the values promoted or demoted by an action. In~\cite{Bench-Capon2017}, arguments for and against actions are used to help agents choose between actions based on their preferences over these values, and the proposed approach aims at justifying norms in general as well as reasoning about when norms should be violated.


In~\cite{Serramia2018}, the authors are interested in the problem of choosing a set of norms that best fit the moral values of a society. In their approach, however, the authors neither provide a formal definition of what values are, nor do they provide a formal model for assessing how much a norm supports (or is aligned with) a value, which they assume is given. While they use preferences to specify which value is preferred to which other value, we use preferences over the states of the world to help us define a single value. As a result, we then assess the alignment of a norm to a value based on whether this norm allows us to move to preferred states with respect to the value in question. 

\label{sec:background}



 


\section{A Formal Approach to Values}\label{sec:values}
\subsection{Values as Preferences}
In this section, we propose a formal model to represent values and ways to compute value aggregation. Values are formally understood as preferences over behaviour or preferences over states of the world; this is in line with the opinion of many philosophers~\cite{ferratermora} and coherent with the model introduced in the background. These preferences usually reflect one's sense of right and wrong, good or bad, and hence, help decide what course of action might be ``better.'' For example, if equality between men and women is one of your values, then you would prefer a state of the world where women and men are paid equally as opposed to one where they are not, and you would support actions leading towards that state of the world. 

We adopt here the traditional view of the world as a labelled transition system~\cite{LTS}: that is, the world is described as a set of states and the different actions allow us to move from one state of the world to another. 

\begin{definition} 
The world is defined as a labelled transition system $(\mathcal{S},\mathcal{A},T)$, where $\mathcal{S}$ is a set of states, $\mathcal{A}$ is a set of actions, and $\mathcal{T}$ is a set of labelled transitions ($T \subseteq \mathcal{S} \times \mathcal{A} \times \mathcal{S}$). For simplification, we use the notation $s \xrightarrow{a} s'$ to describe the transition $(s,a,s') \in T$.
\end{definition}

Values, in a given world, then specify which states of the world are preferred to which other states and to what degree. Note that in our model values are individual and thus have to be associated to a particular agent, human or software. 

\begin{definition}
A value-based preference $\mathsf{Prf}$ over pairs of world states describes how much preferred is one state of the world over another for a given agent with respect to a given value, $\mathsf{Prf}: S \times S \times G \times V \to [-1,1]$, where $G$ is the set of agents and $V$ is the set of values. We use the notation $\mathsf{Prf}^{\alpha}_{v}(s,s')$ to describe how much does $\alpha \in G$ prefer the state of the world $s' \in S$ over $s \in S$ with respect to value $v\in V$. 
\end{definition}

We set the range of preferences to be $[-1,1]$, where a positive number illustrates that $s'$ is more preferred to $s$, a negative number illustrates that it is less preferred, and $0$ illustrates that they are equally preferred. The larger (smaller) the number is, then the more (less) preferred the latter state is to the former. 



\subsection{Aggregation of Value-Based Preferences} 
Although an agent might be capable of assessing what states of the world it prefers with respect to a particular value, trade-offs among values and cumulative effects make that it establishes overall preferences over states of the world for \emph{groups of values}.\footnote{For instance, given $\mathsf{Prf}^{\alpha}_{Security}(s,s') > \mathsf{Prf}^{\alpha}_{Privacy}(s,s')$ what should be $\mathsf{Prf}^{\alpha}_{\{Security, Privacy\}}(s,s')$} Similarly, from a social perspective, the determination of the joint preferences of a group of agents over the states of the world is key to enable their joint planning.\footnote{Given $\mathsf{Prf}^{\alpha}_{Security}(s,s')$ and  $\mathsf{Prf}^{\beta}_{Security}(s,s')$ what should be the value of  $\mathsf{Prf}^{\{\alpha, \beta\}}_{Security}(s,s')$}. Thus a number of aggregation functions can be defined:
\begin{itemize}
    \item {\it One's preference with respect to a set of values:} calculated by aggregating one's preference over each value in that set. 
        \begin{equation}\label{eq:agg1}
            \mathsf{Prf}^{\alpha}_{V} = p(\{\mathsf{Prf}^{\alpha}_{v}\}_{v \in V})
        \end{equation}
    \item {\it A group of people's preference with respect to a given value:} calculated by aggregating each person's preference over that given value.
        \begin{equation}\label{eq:agg2}
            \mathsf{Prf}^{G}_{v} = q(\{\mathsf{Prf}^{\alpha}_{v}\}_{\alpha \in G})
        \end{equation}
    \item {\it A group of people's preference with respect to a set of values:} calculated by aggregating each person's preference over each value in the set.
        \begin{equation}\label{eq:agg3}
            \mathsf{Prf}^{G}_{V} = f(\{\mathsf{Prf}^{\alpha}_{V}\}_{\alpha \in G})
        \end{equation}
           \begin{equation}\label{eq:agg4}
            \mathsf{Prf}^{G}_{V} = g(\{\mathsf{Prf}^{G}_{v}\}_{v \in V})
        \end{equation}
\end{itemize}
Figure~\ref{fig:aggValues} illustrates the relationships between the different aggregation functions. There may be cases that aggregation functions chosen by all agents in a system are coherent with respect to these relationships, for instance a trivial function that works well is the arithmetic average:\footnote{Other average functions would also yield coherence.}

$$\mathsf{Prf}^{\alpha}_V(s,s') = \frac{\displaystyle\sum_{v \in V}\mathsf{Prf}_v^\alpha(s,s')}{|V|}$$
$$\mathsf{Prf}^G_v(s,s') = \frac{\displaystyle\sum_{\alpha \in G}\mathsf{Prf}_v^\alpha(s,s')}{|G|}$$

as 
$$\mathsf{Prf}^G_V(s,s') = \frac{\displaystyle\sum_{\alpha \in G}\mathsf{Prf}_V^\alpha(s,s')}{|G|} = \frac{\displaystyle\sum_{v \in V}\mathsf{Prf}_v^G(s,s')}{|V|}$$

However, in general each agent may choose to combine its preferences over values using a different function/method and thus socially agreeing of a preference over a set of values will depend on the order in which the aggregations are made.

\begin{figure}
\begin{center}
\begin{tikzcd}
\{\mathsf{Prf}_{v}^{\alpha}\}_{\alpha \in G, v \in V} \arrow[r, "p"] \arrow[d, "q"] & \{\mathsf{Prf}_{V}^{\alpha}\}_{\alpha \in G} \arrow[d, "f"] \\
\{\mathsf{Prf}_{v}^G\}_{v \in V} \arrow[r, "g"] & \mathsf{Prf}_{V}^{G}
\end{tikzcd}
\end{center}
\caption{The different value-based preferences and the different aggregation functions}
    \label{fig:aggValues}
\end{figure}

\subsection{Value-Based Preferences based on State Properties}
As illustrated earlier, values specify our preferences over the states of the world. For example, if one values equality between men and women then s/he will most probably prefer a state of the world where men and women are equally paid to another where women are underpaid. 

What distinguishes one state of the world from another are the properties that hold in that state of the world. As such, it is these state properties that influence the preferences between the states of the world. For that, we say values must be related to state properties. 

For example, if one values equality between men and women then the value-based preferences should be influenced by the satisfaction of properties, such as: 1) women and men receiving same salaries, 2) maternity and paternity leaves being equal, etc. Though value-based preferences with respect to equality between men and women should not be influenced by, say, the property of engineer's salaries being in the range [\euro40,000 -- \euro50,000].

Let $\Phi_{v}$ be the set of properties relevant to value $v\in V$. We then say that any value based preference $\mathsf{Prf}_{v}(s,s')$ must be dependent on the satisfaction of $\Phi_{v}$ at states $s$ and $s'$; that is, 
\begin{equation}\label{eq:values-properties}
    \mathsf{Prf}_{v}(s,s') = f(\mathrm{P}(s\models \Phi_{v}), \mathrm{P}(s'\models \Phi_{v}))
\end{equation} 
where $\mathrm{P}(s\models \Phi_{v})$ describes the probability of the satisfaction of the set of properties $\Phi_{v}$ at state $s$, i.e., the degree of satisfaction of $\Phi_{v}$ at state $s$. 

Defining the probability $\mathrm{P}(s\models \Phi_{v})$, as well as defining the function $f$, is outside the scope of this paper and is left for future work; though the example at the end of this paper illustrates how preferences can be based on state properties. 

\section{The Value-Alignment Problem}\label{sec:vap}
The value alignment problem is described, informally, as how much aligned are agents' decisions, and hence actions, with the values that the agents hold dear to them. And since behaviour (decisions and actions) is governed by norms, we describe this alignment as an alignment between the norms that govern behaviour and the values that are held in high regard. 

\begin{figure}[!hb]
    \centering
    \subfloat[A world with no taxes\label{fig:nw1}]{\includegraphics[width=0.4\textwidth]{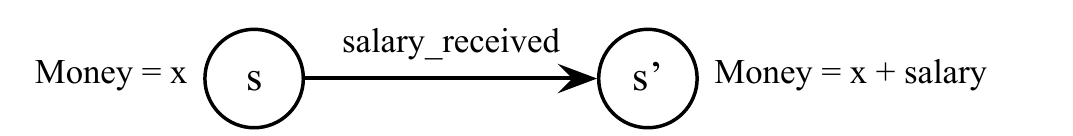}}\hfill
    \subfloat[A world with 20\% taxes\label{fig:nw2}]{\includegraphics[width=0.4\textwidth]{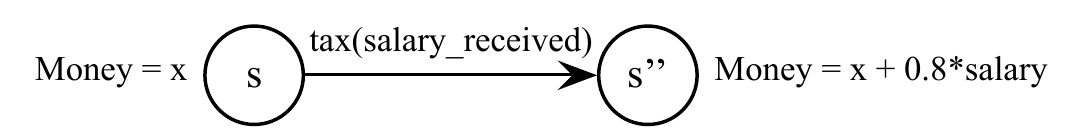}}
    \caption{Applying a norm to a given world alters the transitions and their resulting states}
    \label{fig:normative–world}
\end{figure}

We understand norms as rules that govern behaviour. We say a norm $n \in N$ (where $N$ is a set of norms) is a logical formula that describes the conditions under which a certain action can/cannot be performed along with the post-conditions of that action. When a set of norms $N$ is applied to a world $(\mathcal{S},\mathcal{A},T)$, the world is modified by he norms in $N$, resulting in a new world $(\mathcal{S},\mathcal{A},N,T_{N})$, which we refer to as a normative world. For example, in a world where people do not get taxed, your money increases by the amount of your salary when your salary is paid (see the action `salary\_received' and the new state $s'$ of Figure~\ref{fig:nw1}, where $Money$ describes how much money one has at a given state). However, the norm of another country that introduces a 20\% tax on your income essentially modifies the action of receiving your salary (`salary\_received') by applying the tax, and hence, resulting in a transition to a new state where your income is deducted by 20\% (state $s''$ in Figure~\ref{fig:nw2}). 

\begin{definition}
A normative world $(\mathcal{S}_{N},\mathcal{A},N,T_{N})$ describes the world $(\mathcal{S},\mathcal{A},T)$ where the set of norms $N$ have been applied to the transitions in $T$, resulting in possibly new transitions and states. 
\end{definition}

How much a given norm $n \in N$ is aligned to a given value $v \in V$ with respect to a world $(\mathcal{S},\mathcal{A},T)$ then depends on whether applying norm $n$ would result in new transitions ($T_{N}$) that would move us to preferred (and possibly new) states or not. To be able to calculate this, we will need to have a list of the different paths in a given world, which we define accordingly.

\begin{definition}
A path $p$ in a world $(\mathcal{S},\mathcal{A},T)$ is a sequence of transitions in $T$:
$[s \xrightarrow{\alpha} s', \ldots, s'' \xrightarrow{\beta} s''']$,
such that $p_F[i] = p_I[i+1]$, where $p_I[i]$ represents the initial state of the $i$’th transition in $p$ and $p_F[i]$ represents the final state of the $i$’th transition in $p$. In other words, the final state of every transition equals the initial state of the following transition.
\end{definition}

Alignment is then defined as follows.
\begin{definition}
The degree of alignment of a norm $n\in N$ with a value $v\in V$ with respect to a world $(\mathcal{S},\mathcal{A},T)$ for a given agent $\alpha$ is defined through the accumulated preferences in the resulting normative world that applies norm $n$ (that is, the world $(\mathcal{S}_{N},\mathcal{A},N,T_{N})$), which is specified as: 
$$\mathsf{Algn}_{n,v}^{\alpha}(\mathcal{S},\mathcal{A},T) = \displaystyle\frac{\displaystyle\sum_{p\in paths}\displaystyle\sum_{d\in [1,length(p)]} \mathsf{Prf}_{v}^{\alpha}(p_I[d],p_F[d])}{\displaystyle\sum_{p\in paths}length(p)}$$
where $paths$ is the set of all paths in world $(\mathcal{S}_{\{n\}},\mathcal{A},\{n\},T_{\{n\}})$, and $length(p)$ describes the  length of a path $p \in paths$.
\end{definition}
In other words, considering all possible paths in the normative world that applies norm $n$, $(\mathcal{S}_{\{n\}},\mathcal{A},\{n\},T_{\{n\}})$, we calculate the average change in preferences for each transition of those paths. Of course, our proposal for calculating alignment is an initial proposal that gives equal weight to all paths and all transitions of a path. Alternative approaches may also be considered, which we leave for future work. For example, the variance in the preferences might be indicative, say if one prefers steady increase in preferences over fluctuating preferences. Furthermore, if more knowledge about this world is available, such as the probability of transitions, then one can take this knowledge into consideration and can give less probable paths (or transitions) less weight than others. Similarly, if knowledge about which states are more important (regardless of whether they are less or more preferable), the preference of these states can be given more weight. One, for example, may want to give more weight to states that are in the distant future than those that are in the immediate future as the distant future might be more important.

Note that as preferences are subjective with respect to a given agent $\alpha$, the alignment of a norm to a value is then also subjective with respect to the same agent $\alpha$. Also note that alignment is described by positive numbers whereas misalignment by negative numbers, and the higher the number, then the more aligned is the norm with the value in question, and vice versa. 

Of course, the definition above requires calculating the preferences between states for all possible transitions in a given world $(\mathcal{S}_{\{n\}},\mathcal{A},\{n\},T_{\{n\}})$. This is not an efficient approach. As such, we propose to use the Monte Carlo sampling method to randomly select some of the paths in this world. Furthermore, we also suggest to restrict the length of these paths, which is useful especially in infinite state spaces. We say let $l$ describe the length of paths, and $x$ the number of sampled paths. Then, alignment can be calculated as follows:

\begin{equation}\label{eq:align}
\mathsf{Algn}_{n,v}^{\alpha}(\mathcal{S},\mathcal{A},T) = \displaystyle\frac{\displaystyle\sum_{p\in paths'} \sum_{d\in [1,l]} \mathsf{Prf}_{v}^{\alpha}(p_I[d],p_F[d])}{x*l}
\end{equation}
where $paths'$ is a set of $x$ randomly selected paths of length $l$ in the normative world $(\mathcal{S}_{\{n\}},\mathcal{A},\{n\},T_{\{n\}})$.


As with preferences, alignment can be calculated for sets of values, sets of norms, and/or sets of agents, so we can calculate $\mathsf{Algn}_{n,V}^{\alpha}(\mathcal{S},\mathcal{A},T)$, $\mathsf{Algn}_{N,V}^{\alpha}(\mathcal{S},\mathcal{A},T)$, or $\mathsf{Algn}_{N,V}^{G}(\mathcal{S},\mathcal{A},T)$, and so on.

%



In addition to alignment, we also define the relative alignment of norm $n_{1}$ with respect to $n_{2}$ for a given value $v$ accordingly. 
\begin{definition}
The relative alignment of norm $n_{1}$ with respect to $n_{2}$ for a given value $v$ in a given world $(\mathcal{S},\mathcal{A},T)$ describes how much more $n_{1}$ is aligned with $v$ than $n_{2}$ is aligned with $v$, and it is specified as:
$$\mathsf{RAlgn}_{n_{1}/n_{2},v}^{\alpha}(\mathcal{S},\mathcal{A},T) = \mathsf{Algn}_{n_{1},V}^{\alpha}(\mathcal{S},\mathcal{A},T) - \mathsf{Algn}_{n_{2},V}^{\alpha}(\mathcal{S},\mathcal{A},T)$$
\end{definition}
Where positive numbers imply $n_{1}$ is more aligned than $n_{2}$ with respect to $v$, and negative numbers imply the opposite holds.

Again, as above, the relative alignment can be calculated for sets of values, sets of norms, and/or sets of agents, as needed.

Last, but not least, we note that computing alignment is based on the assumption that all transitions in a given world are given the same weight. In other words, we assume that transitions are equiprobable to occur. Of course, in reality this is not true due to a couple of reasons. First and foremost, the probability of reaching a given state is not the same for all states, as the norms (or the rules that govern behaviour) might result in having one state more (or less) probable to reach than others. Second, the probability of agents choosing one action over another cannot be predicted and decisions are usually not equiprobable. However, for simplicity, this paper assumes all transitions are equiprobable.

\section{Example}\label{sec:example}
Let's illustrate the concept of value alignment with a simple example: the traditional Prisoners' Dilemma whose payouts are presented in Table~\ref{tbl:pd}. The game is played repeatedly. Although traditionally this example assumes  self interest and rationality of the players, we'll see that within our framework we can tweak it (via norms) so that values other than selfishness can be accommodated. 
\begin{table}[h]
     \caption{The Prisoner’s Dilemma.}\label{tbl:pd} 
     \begin{center}
     \begin{tabular}{|c | c | c |}
     \cline{2-3}
     \multicolumn{1}{c|}{}& {\bf $\beta$ Co-operates} & {\bf $\beta$ Defects} \\\hline%
     {\bf $\alpha$ Co-operates} & 6,6 & 0,9 \\\hline
     {\bf $\alpha$ Defects} & 9,0 & 3,3 \\\hline
     \end{tabular}
     \end{center}
\end{table}

The states of the world are described through the accumulated gain of each agent, which we will represent as $(x,y)$, where $x$ stands for $\alpha$'s accumulated gain and $y$ for $\beta$'s. Every time a game is played, extra gains are accumulated to make the state change from one state ($s$) to another ($s'$), with their properties changing accordingly from $(x,y)$ to $(x',y')$, with $x' \geq x$ and $y'\geq y$.

\subsection{Value-Based Preferences}
In this example, we will consider the value \emph{equality}. Equality might mean different things for different people, and it is usually valued differently by different people. 
We present four different functions that could be used to define preferences with respect to the value `equality'. Note that preferences are defined in terms of the properties specifying the accumulated gains at each state.
\begin{itemize}
    \item States with higher equality in accumulated gain are preferred:
    \begin{equation}\label{eqpodemos}
        \mathsf{Prf}(s,s') =    \frac{|x-y|}{\max{\{x,y\}}}-\frac{|x'-y'|}{\max{\{x',y'\}}} 
    \end{equation}
    \item States with higher equality in accumulated gain are preferred only if my personal gain is not lower:
    \begin{equation}\label{eqciudadanos}
        \mathsf{Prf}(s,s') = \left(1-\frac{|y'-x'|}{\max{\{x',y'\}}}\right)\cdot \frac{x'-x}{\max{\{x',x\}}}    
    \end{equation}
    \item States with higher personal gain are preferred only if equality is not lower:
    \begin{equation}\label{grpodemos}
        \mathsf{Prf}(s,s') =  \frac{x'-x}{2(\max{\{x',x\})}}-\frac{y'-y}{2(\max{\{y',y\})}}
    \end{equation}
    \item States with higher personal gain are preferred, regardless of equality:
    \begin{equation}\label{grciudadanos}
        \mathsf{Prf}(s,s') =  \frac{x'-x}{\max{\{x',x\}}} 
    \end{equation}
\end{itemize}

\subsection{Norms}

We will define three examples of norms that introduce taxes over the gains of agents playing the prisoner's dilemma:
\begin{itemize}
    \item [{\it No taxing}.] $n_0$: No taxes are to be payed.
    \item [{\it Incremental taxing}.] $n_1$: Taxes will be paid as follows: no taxes to be paid when the gain is 0 or 3, 3 to be paid as taxes when the gain is 6, and 5 to be paid as taxes when the gain is 9.
    \item [{\it Fixed taxing}.] $n_2$: $1/3$ of the gains of each game is to be paid as taxes.
\end{itemize}

\begin{figure}[t!]
    \centering
    \subfloat[$n_0$ applied\label{fig:n0}]{\includegraphics[width=0.3\textwidth]{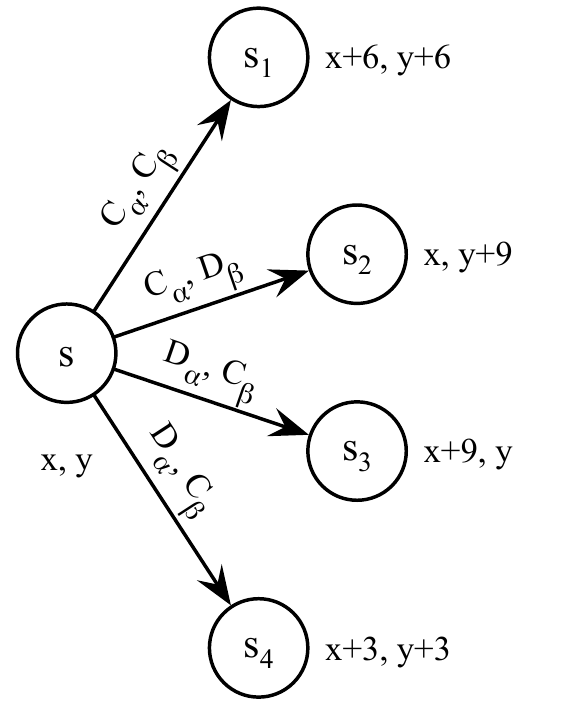}}\hfill
    \subfloat[$n_1$ applied\label{fig:n1}]{\includegraphics[width=0.3\textwidth]{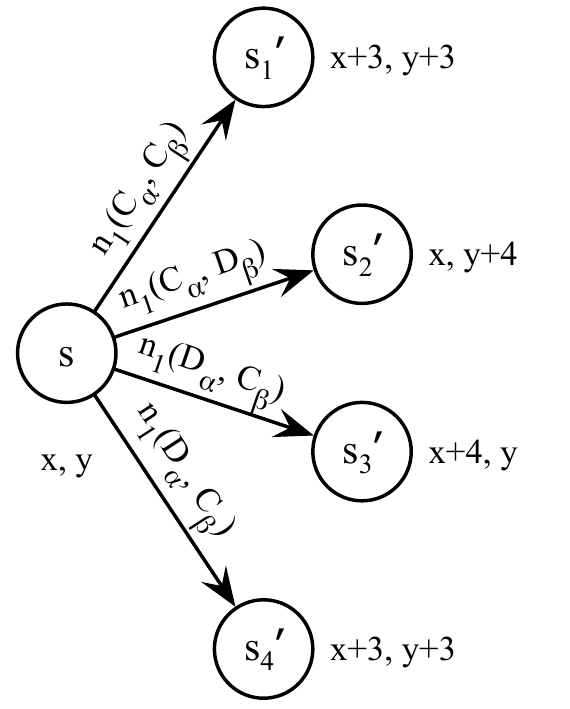}}\hfill
    \subfloat[$n_2$ applied\label{fig:n2}]{\includegraphics[width=0.3\textwidth]{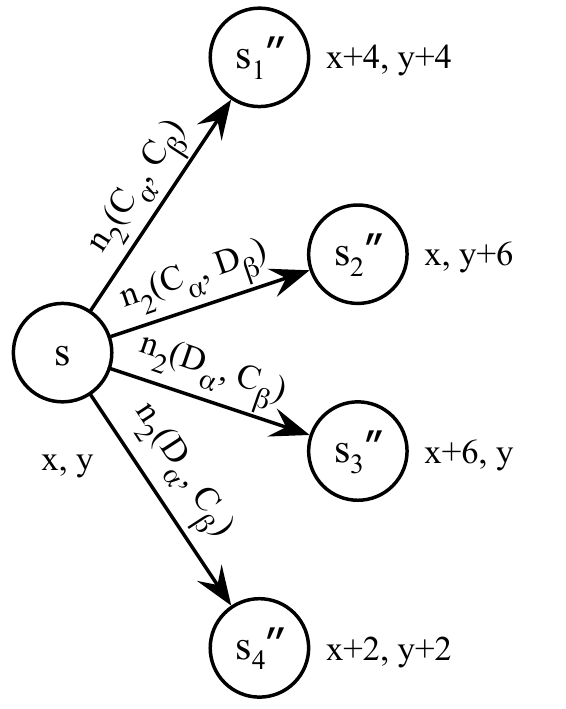}}
    \caption{Applying norms alters the world}
    \label{fig:normative–worldEG}
\end{figure}

Norms $n_1$ and $n_2$ modify the world that applies $n_0$ (Figure~\ref{fig:n0}), as illustrated by Figures~\ref{fig:n1} and~\ref{fig:n2}, respectively.

\subsection{Value Alignment}

The question now is which norms are better aligned with an agent's interpretation of the value `equality' (where different interpretations are possible, following the different Equations~\ref{eqpodemos}--\ref{grciudadanos}). 
The iterated prisoner's dilemma outcome depends on the strategies played by the agents. As this aspect is not the focus of this paper, we will assume that agents will choose their actions randomly. In columns two and three of Table~\ref{results}, you can find the set of actions from which each agent chooses their actions. In column one, the preference modeling value equality is found, and in the last column, the relative alignment of the three norms for that preference are presented.

\begin{table}[ht]
    \begin{center}
        \caption{The relative alignment of norms $n_0$, $n_1$ and $n_2$ with respect to different definitions of the value equality that $\alpha$ may adopt and different sets of actions that $\alpha$ and $\beta$ may choose from, randomly. Sampling is set to 20,000 and the game is played 10 times (that is, $x=20,000$ and $l=10$ in Equation~\ref{eq:align}).}\label{results}
    \begin{tabular}{c|c|c|c|c|}
    \cline{2-5}
    \multicolumn{1}{c|}{} & $\mathsf{Prf}_{eq}^{\alpha}$ &  $\alpha$'s actions & $\beta$'s actions & Relative Alignment \\ \cline{2-5}\cline{2-5}
    1 & Eq.~\ref{eqpodemos} &\{c\}  & \{c,d\}& $n_1 \succ n_0 \sim n_2$ \\ \cline{2-5}
    2 & Eq.~\ref{eqciudadanos} &\{c\}  & \{c,d\}& $n_0 \sim n_1 \sim n_2$ \\ \cline{2-5}
    3 & Eq.~\ref{grpodemos} &\{c\}  & \{c,d\}& $n_0 \sim n_1 \sim n_2$ \\ \cline{2-5}
    4 & Eq.~\ref{grciudadanos} &\{c\}  & \{c,d\}& $n_0 \succ n_2 \succ n_1$ \\ \cline{2-5}
    5 & Eq.~\ref{eqpodemos} &\{d\}  & \{c,d\}& $n_1 \succ n_0 \sim n_2$ \\ \cline{2-5}
    6 & Eq.~\ref{eqciudadanos} &\{d\}  & \{c,d\}& $n_0 \sim n_1 \sim n_2$ \\ \cline{2-5}
    7 & Eq.~\ref{grpodemos} &\{d\}  & \{c,d\}& $n_0 \sim n_1 \sim n_2$ \\ \cline{2-5}
    8 & Eq.~\ref{grciudadanos} &\{d\}  & \{c,d\}&  $n_0 \sim n_1 \sim\succ n_2$\\ \cline{2-5}
    9 & Eq.~\ref{eqpodemos} &\{c,d\}  & \{c\}&  $n_1 \succ n_0 \sim n_2$ \\ \cline{2-5}
    10 & Eq.~\ref{eqciudadanos} &\{c,d\}  & \{c\}&  $n_0 \sim n_1 \sim n_2$ \\ \cline{2-5}
    11 & Eq.~\ref{grpodemos} &\{c,d\}  & \{c\}&  $n_0 \sim n_1 \sim n_2$\\ \cline{2-5}
    12 & Eq.~\ref{grciudadanos} &\{c,d\}  & \{c\}&  $n_0 \sim n_1 \sim n_2$ \\ \cline{2-5}
    13 & Eq.~\ref{eqpodemos} &\{c,d\}  & \{d\}&  $n_1 \succ n_0 \sim n_2$ \\ \cline{2-5}
    14 & Eq.~\ref{eqciudadanos} &\{c,d\}  & \{d\}& $n_1 \succ n_0 \sim n_2$ \\ \cline{2-5}
    15 & Eq.~\ref{grpodemos} &\{c,d\}  & \{d\}&  $n_1 \succ n_0 \sim n_2$ \\ \cline{2-5}
    16 & Eq.~\ref{grciudadanos} &\{c,d\}  & \{d\}&  $n_0 \sim n_1 \succ n_2$  \\ \cline{2-5}
    17 & Any & \{c,d\}  & \{c,d\}&  $n_0 \sim n_1 \sim n_2$  \\ \cline{2-5}
   \end{tabular}
    \end{center}
\end{table}

The results of Table~\ref{results} illustrate the following. No matter the actions chosen by $\alpha$ or $\beta$, the norm better aligned with a strong support to equality (specified through Equation~\ref{eqpodemos}, see lines 1, 5, 9, 13) is incremental taxing ($n_1$). Moderate supporters of equality (specified through Equations~\ref{eqciudadanos} and \ref{grpodemos}) have no norm specially well aligned (lines 2, 3, 6, 7, 10, and 11), except when the gains of $\beta$ are higher (by always choosing to defect: choosing action $d$) in which case they consider incremental taxing better aligned ($n_1$) (lines 14, 15). Finally, when there is a random selection over $[c,d]$ by both players (line 17) leading then to an egalitarian society, there is no preferred norm, as none of them increase inequality over an egalitarian society.

\section{Conclusions and Suggested Work}\label{sec:challenges}
This paper has provided a formal model that defines values as preferences over states of the world, and value-alignment through the increase/decrease of preferences in a given world. A computational model has been presented for calculating the degree of value-based preferences, the degree of the alignment of a norm to a value in a given world, as well as the relative alignment of one norm with respect to another for a given value in a given world. 

Future work should help define the different aggregation functions for values (Equations~\ref{eq:agg1}--\ref{eq:agg4}). For example, how do social values arise from individual values? 
Future work should also help define the $f$ function and the probability $\mathrm{P}(s\models \Phi_{v})$ of Equation~\ref{eq:values-properties}, which describe how preferences are generated. Last, but not least, future work should study the impact of the assumption made that all transitions are considered equiprobable. 

Nevertheless, with this initial formal model for values and value-alignment, we can now formalise questions that can help us study agent societies. 
Given a set of norms $N$, a set of values $V$ and set of agents $G$ such that $\mathsf{Algn}_{n,v}^{\alpha}$ and $\mathsf{Prf}_{v}^{\alpha}$ are known for all $n \in N$, $v \in V$, and $\alpha \in G$, then we can formalise the following questions:
\begin{question}
    What is the subset of norms $N^* \subseteq N$ with optimal alignment for group $G$? That is, how to compute:
    $$ N^* = \arg \max_{N' \subseteq N} \mathsf{Algn}_{N',V}^{G}$$
\end{question}
\begin{question}
   What is the subset of agents $G^* \subseteq G$ better aligned with norms $N$? That is, how to compute:
    $$ G^* = \arg \max_{G' \subseteq G} \mathsf{Algn}_{N,V}^{G'}$$
\end{question}
\begin{question}
What is the optimal social preference aggregation function? That is, how to compute:
    $$ f^* = \arg \max_{f \in F} \mathsf{Algn}_{N,V}^{G'}(f\{\mathsf{Prf}_{V}^{\alpha}\}_{\alpha \in G})$$
\end{question}

\section{Acknowledgments}
This work has been supported by the Catalan funded AppPhil project (funded by RecerCaixa 2017), the Spanish funded CIMBVAL project (funded by the Spanish government, project \# TIN2017-89758-R), and the EU funded WeNet project (funded under the H2020 FET Proactive 2018 call, project \# 823783). 

\bibliographystyle{splncs04}
\bibliography{references}
\end{document}